\documentclass[10pt,twocolumn,letterpaper]{article}

\usepackage{cvpr}
\usepackage{times}
\usepackage{epsfig}
\usepackage{graphicx}
\usepackage{subfigure}
\usepackage{amsmath}
\usepackage{amssymb}
\usepackage{algorithm}
\usepackage{algorithmic}
\usepackage{array}
\usepackage{bm}
\usepackage{booktabs}
\usepackage{tabularx}

\usepackage[breaklinks=true,bookmarks=false]{hyperref}

\cvprfinalcopy 

\def\cvprPaperID{****} 

\begin{document}

\title{RepNAS: Searching for Efficient Re-parameterizing Blocks}

\author{Mingyang Zhang, Xinyi Yu, Jingtao Rong, Linlin Ou\\
College of Information Engineering, Zhejiang University of Technology\\
Hang Zhou, People’s Republic of China\\
{\tt\small linlinou@zjut.edu.cn}
}

\maketitle

\begin{abstract}
In the past years, significant improvements in the field of neural architecture search(NAS) have been made. However, it is still challenging to search for efficient networks due to the gap between the searched constraint and real inference time exists. To search for a high-performance network with low inference time, several previous works set a computational complexity constraint for the search algorithm. However, many factors affect the speed of inference(e.g., FLOPs, MACs). The correlation between a single indicator and the latency is not strong. Currently, some re-parameterization(Rep) techniques are proposed to convert multi-branch to single-path architecture which is inference-friendly. Nevertheless, multi-branch architectures are still human-defined and inefficient. In this work, we propose a new search space that is suitable for structural re-parameterization techniques. RepNAS, a one-stage NAS approach, is present to efficiently search the optimal diverse branch block(ODBB) for each layer under the branch number constraint. Our experimental results show the searched ODBB can easily surpass the manual diverse branch block(DBB) with efficient training.
\end{abstract}
\begin{figure}[t]
    \centering
    \includegraphics[width=3.5in]{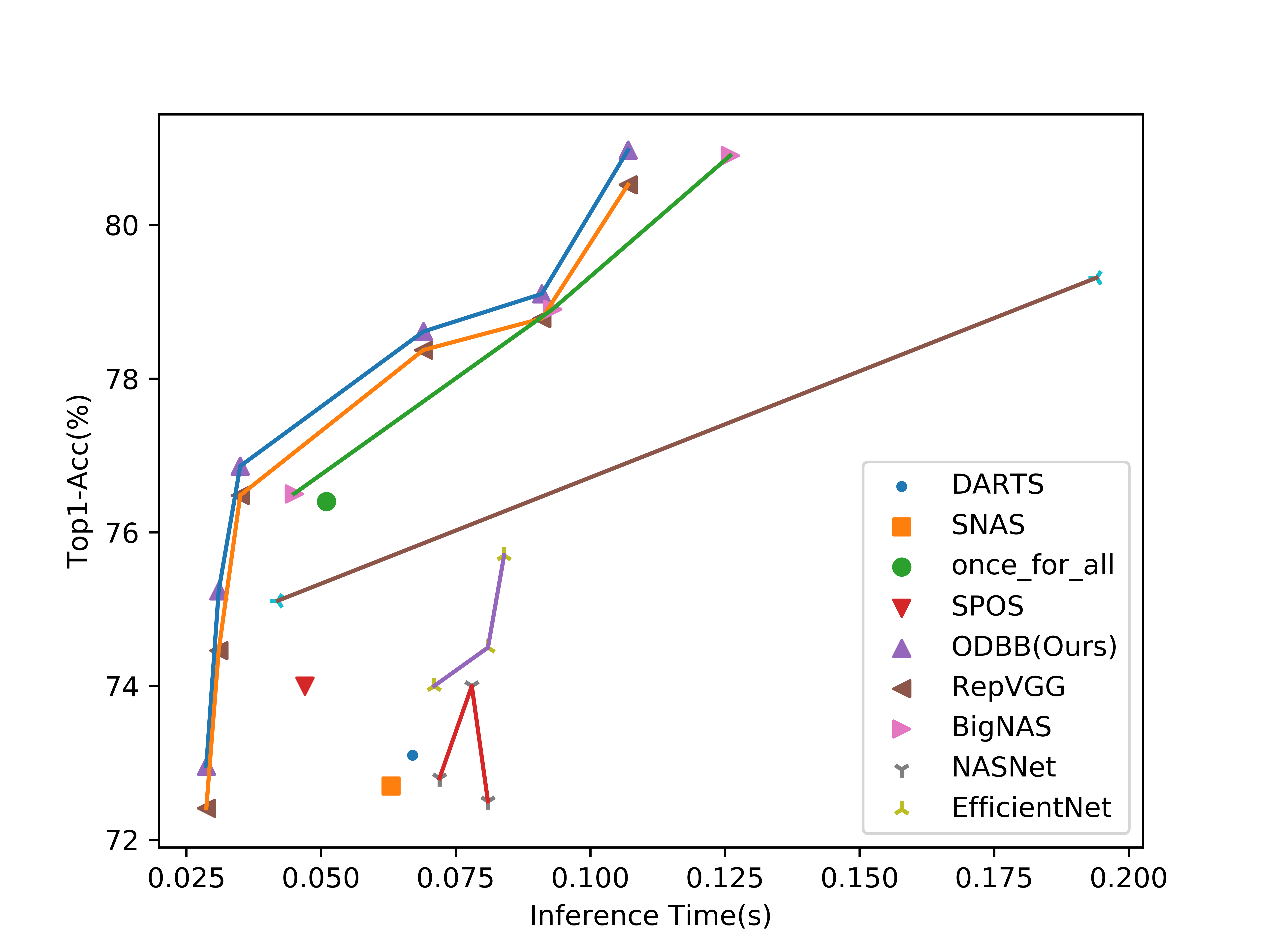}
    \caption{The comparison between ODBBs and other architecture-searched models. The inference time is tested on a Tesla V100 GPU with a batch size of 128, full precision(fp32).}
    \label{fig:performance}
\end{figure}
\section{Introduction}

Designing low-latency neural networks is critical to the application on mobile devices, e.g., mobile phones, security cameras, augmented reality glasses, self-driving cars, and many others. Neural architecture search(NAS) is expected to automatically search a neural network where the performance surpasses the human-designed one under the most common constraints, such as latency, FLOPs and parameter number. One of the key factors for the success of NAS is the artificially designed search space. Many previous NAS methods are designed to search on DARTS space\cite{liu2018darts,xie2018snas,real2019regularized} and MobileNet-like space\cite{hu2020dsnas,tan2019efficientnet,yu2020bignas}. DARTS space contains a multi-branch topology and a combination of various paths with different operations and complexities. The multi-branch design can enrich the feature space and improve the performance\cite{szegedy2017inception}. Multi-branch structure is also used in some neural networks designed by human before, e.g., the Inception models\cite{szegedy2017inception} and DenseNet models\cite{iandola2014densenet}. However, these multi-branch architectures result in a longer inference time that is unfriendly for real tasks. Therefore, many applicable NAS methods are adopted to MobileNet-like space by keeping only two branches(skip connection and convolutional operation) for searched networks to realize a tradeoff between the inference time and the performance. 

To maintain the inference speed of a single branch while retaining the advantages of multiple branches, several Rep techniques\cite{ding2021repvgg,ding2019acnet,ding2021diverse,ding2021repmlp,ding2020lossless} have been introduced. Rep techniques can retain multi-branch topology at training time while keeping a single path at running time, and thus can realize a balance of accuracy and efficiency. The multi-branch can be fused to one path because of linear operation, for example, a 3x3 Conv and a 1x1 Conv can be replaced by a 3x3 Conv by padding the 1x1 kernel to 3x3 and element-wise adding with the 3x3 Conv. It is worthy to note that the fused network has a lower inference time while keeping almost the same accuracy compared with the multi-branch network. 

However, prior works that utilize Rep techniques to train models have some limitations. \textbf{1)} The branch number and branch types of each block. The multi-branch training requires multiple memory(increasing linearly with the branches number) to save the middle feature representations. Therefore, the branch number and branch types of each block are manually fixed because of memory constraints. For instance, the RepVGG block only contains a 1x1 Conv, a 3x3 Conv and a skip connection while the ACB has 1x3 Conv and 3x1 Conv. \textbf{2)} The role of each branch is unclear, which means some branches may counterproductive. Ding\cite{ding2021diverse} experimented with various micro Diverse Branch Block(DBB) structures to explore which structure works better. However, on one hand, the manual micro DBB(all blocks are the same) is always suboptimal. Many NAS work\cite{guo2020single,yu2020bignas,hu2020dsnas} revealed that the optimal structure of blocks is different from each other. On the other hand, it is infeasible to design macro DBB of which search space approximately reaches $1.5 \times 10^{36}$(Assuming there have 30 blocks and each block has 4 branches). 

To address the limitation \textbf{1)}, we devise a multi-branch search space, called Rep search space. Unlike previous works that use Rep techniques, each block can preserve an arbitrary number of branches and all block architectures are independent in this paper. Facing increasing memory, the total branch number of the model can be flexibly adjusted. To address the limitation \textbf{2)}, a gradient-based NAS method, RepNAS, is proposed to automatically search the optimal branch for each block without parameter retraining. Compared with previous gradient-based NAS methods\cite{liu2018darts,xie2018snas} that only learn the importance in one edge, RepNAS learns the net-wise importance of each branch. Moreover, RepNAS can be used under low GPU memory conditions by setting the branch number constraint. In each training iteration, the branches of low importance will be sequentially pruned until the memory constraint is met.
The importance of branches is updated in the same round of back-propagation as gradients to neural parameters. As the training progresses, the importance of each branch is estimated accurately, meanwhile, the redundant branches hardly participate in the training. Once the training process
finished, the optimized DDB structure is obtained with optimized network parameters. 

To summarize, our main comtributions are as follow:
\begin{itemize}
  \item A Rep search space is proposed in this paper, which allows the searched model to preserve arbitrary branches in training and can be fused into one path in inference. To our best knowledge, it is the first time that Rep techniques can be used to NAS.
  \item To fit the new search space, RepNAS is presented to automatically one-stage search the optimal DBB. The searched model can be converted to a single-path model and directly deployed without time-consuming retraining.
  \item Extensive experiments on models with various sizes demonstrate that the searched ODBB outperform both the human-designed DDB and NAS models under similar inference time. 
\end{itemize}

\section{Related Work}
\label{sec:related-word}
\subsection{Structural Re-parameterization}
\label{subsec:re-param}
Structural re-parameterization techniques have been widely used to improve model performance by injecting several branches in training and fusing these branches in inference. RepVGG\cite{ding2021repvgg} simply inserted a $1 \times 1 Conv$ and a residual connection into a $3 \times 3 Conv$, which makes the performance of VGG-like networks have a huge improvement. A concurrent work, DBB\cite{ding2021diverse}, summarized more structural re-parameterization techniques and proposed a Diverse Branch Block which can be inserted into any convolutional network. Here, we list the existing techniques as follows

\textbf{a Conv for Conv-BN.} BN\cite{ioffe2015batch} can be fused into its preceding Conv parameters $F$ and $b$ for inference. The $j$th fused channel parameters $F^{'}_{j,:,:,:}$ and $b^{'}_{j}$ can be formulated as
\begin{equation}
\label{trans1}
  F^{'}_{j,:,:,:} = \frac{\gamma_j}{\sigma_j}F_{j,:,:,:},\ b^{'}_{j} = - \frac{\mu_j \gamma_j}{\sigma_j} + \beta_j
\end{equation}
where $\gamma$, $\sigma$ and $\beta$ denotes the learned scaling factor and bias term of BN, respectively.

\textbf{A Conv for branch Conv addition.} Convs with different kernel size in different branches can be fused into one Conv(without nonlinear activation). The kernel size of each Conv should be padded to the maximum of them. And then, they can be merged into a single Conv by
\begin{equation}
\label{trans2}
  F^{'} = F^{1} + F^{2} + ... + F^{N},\ b^{'} = b^{1} + b^{2} + ... + b^{N}
\end{equation}
where $N$ denotes the branch number.

\textbf{A Conv for sequential Convs.} A sequence of Convs with $1\times 1$ - $k \times k$ can be merged into one single $K \times K$ Conv. 

\begin{equation}
\label{trans3}
  F^{'} = F^{2} * TRANS(F^{1}),\\ b^{'}_{j} = \sum_{d=1}^{D}\sum_{u=1}^{K}\sum_{v=1}^{K}b^{1}_{d}F^{2}_{j,d,u,v} + b^{2}
\end{equation}
where $TRANS$ represents transpose. $1$ and $2$ denote $1\times 1$ Conv and $3\times 3$ Conv, respectively.

\textbf{A Conv for average pooling.} A $k\times k$ average pooling can be viewed as a special $k\times k$ Conv. Its kernel parameter $F^{'}$ is 
\begin{equation}
\label{trans4}
  F^{'} = \frac{1}{k^2}I_{k\times k}
\end{equation}
where $I_{k\times k}$ is $k\times k$ identity matrix. 

However, the architecture of each block is task-specific. For simple tasks or tiny-scale datasets, too many branches lead to overfitting. Besides, the output of each branch needs to be saved in GPU memory for backward. Limited GPU memory prevents the structural Re-parameterization techniques from the application.
\begin{figure*}[t]
    \centering
    \includegraphics[width=7in]{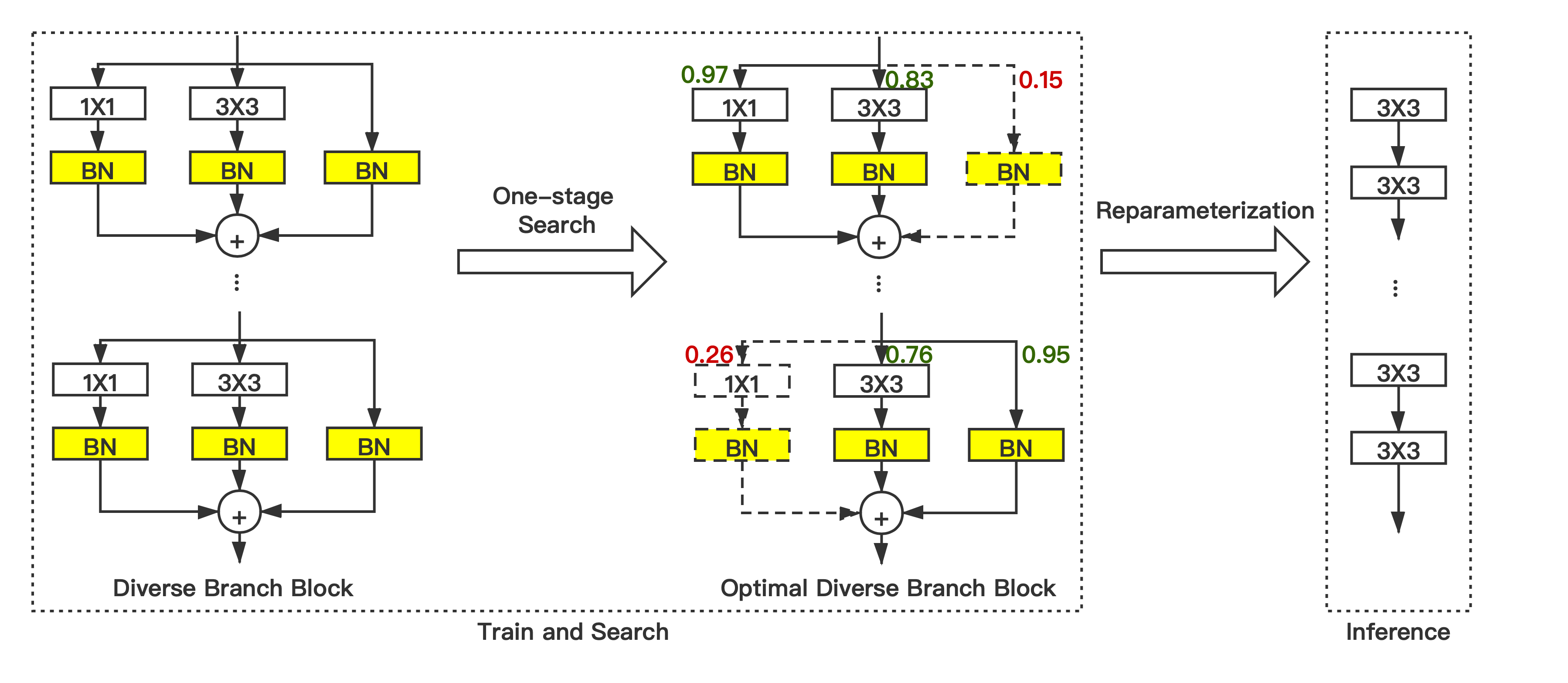}
    \caption{The flow diagram of our proposed approach. Our approach automatically designs architectures of ODBBs at the macro level through pruning the branches with low importance(red number). The architecture searching and network training is finished at the same time, which is contained in the left dash box. The multi-branch architecture of each block will be converted to one path for acceleration(right dash box).}
    \label{fig:overview}
\end{figure*}
\subsection{Neural Achitecture Search}
\label{subsec:NAS}
The purpose of neural network structure search is to automatically find the optimal network structure with reinforcement learning(RL)\cite{zoph2016neural,zoph2018learning}, evolutionary algorithm(EA)\cite{real2019regularized,guo2020single,yu2020bignas}, gradient\cite{liu2018darts,xie2018snas} methods. RL-based and EA-based methods need to evaluate each sampled network by retraining them on the dataset, which are time-consuming. The gradient-based method can simultaneously train and search the optimal subnet by assigning a learnable weight to each candidate operation. However, the gradient-based approach causes incorrect ranking results. A subnet has top proxy accuracy while performing not as expected. Moreover, since gradient-based approaches need more memory for training, they cannot be applied to the large-scale dataset. To address the above problem, some one-stage NAS methods\cite{xie2018snas,hu2020dsnas,dong2019searching} are proposed to simultaneously optimize the architecture and parameters. Once the supernet training is complete, the top-performing subnet is also given without retraining. 

\section{Re-parameterization Neural Achitecture Search}
\label{sec:repnas}
In this section, we first present an overview of our proposed approach for searching optimal diverse branch blocks and discuss the difference with other NAS work. We then propose a new search space based on some Rep techniques mentioned in Eq. \ref{trans1}-\ref{trans4}. Afterward, the RepNAS approach is presented to fit the proposed search space. 
\subsection{Overview}
\label{overview}
The goal of Rep techniques is to improve the training effect of a CNN by inserting various branches with different kernel size. The inserted branches can be linearly fused into the original convolutional branch after training, such that no extra computational complexity(FLOPs, latency, memory footprint or model size) is subjoined. However, training with various branches costs a large GPU memory consumption, and it is hard to optimize a network with too many branches. The core idea of the proposed method is to prune out some branches across different blocks in a differentiable way, which is shown in Figure \ref{fig:overview}. It has two essential steps:

\quad (1) Given a CNN architecture(e.g., MobileNets, VGGs), we net-wisely insert several linear operations into original convolutional operations as its branches. For each branch, a learnable architecture parameter that represents the importance is set. During the training, we optimize both architecture parameters and network parameters by discretely sampling branches, simultaneously. Once training finishes, we can obtain a pruned architecture with optimized network parameters. 

\quad (2) In inference, the rest of the branches can be directly fused into the original convolutional operations, such that the multi-branch architecture can be converted to the single-path architecture without a performance drop. No extra finetune is required in this step.

Compared with many NAS work, cumbersome architectures with various branches and skip connections are no longer an obstacle to application in RepNAS. In contrast to prior structural re-parameterization work, the architectures of blocks in each layer can be automatically designed without any extra time consumption. The whole optimization is in one stage.

\subsection{Rep Search Space Design}
\label{subsec:multi-branch search space}
NAS methods are usually designed to search for the optimal subnetwork on DARTS space or MobileNet-like space. The former contains multi-branch architecture, which makes the searched network difficult to apply becausfunne of the large inference time. The latter which refers to the expert experience in designing mobile networks includes efficient networks, however, multi-branch architecture into no consideration. Many human-designed networks\cite{szegedy2017inception,iandola2014densenet} demonstrate multi-branch architecture can improve model performance by enhancing the feature representation from the various scale. To combine the advantages of multi-branch and single-path architecture, a more flexible search space is proposed based on some current structural re-parameterization work. Shown in Figure \ref{fig:search_space}, each block contains 7 branches($1\times 1$,$K\times K$, $1\times 1 - K \times K$, $1\times 1 - AVG$, $1\times K$, $K \times 1$ and $skip connect$). Different from previous search space, we release the heuristic constraints to offer higher flexibility to the final architecture, which means each block can preserve arbitrary branches and all block architectures are independent. It is worthy to note that the multi-branch will be fused to a single-path after searching, thus, have no impact on inference.

The new search space reaches approximately $2.29\times 10^{105}$ architectures. Compared with either micro($1.1 \times 10^{18}$) or macro search space($1.9\times 10^{93})$, the proposed search space has greater potential to offer better architectures and evaluate the effectiveness of NAS algorithms. However, such an enlarged search space brings challenges for preceding NAS approaches:1) incorrect architecture rattings\cite{chu2019fairnas}. 2) large memory cost because of multi-branch\cite{chen2019progressive}.

\subsection{Weight Sharing for Network Parameters}
Many previous NAS methods share weights across architectures during supernet training while decoupling the weights of different branches at the same block. However, this strategy does not work in the proposed Rep search space because of a surging number of subnets. In each training iteration, only a few branches can be sampled which results in the weights being updated by limited times. Therefore, the performances of sampled subnets grow slowly. This limits the learning of architecture parameter $\alpha$. 
\begin{figure*}[t]
    \centering
    \includegraphics[width=7in]{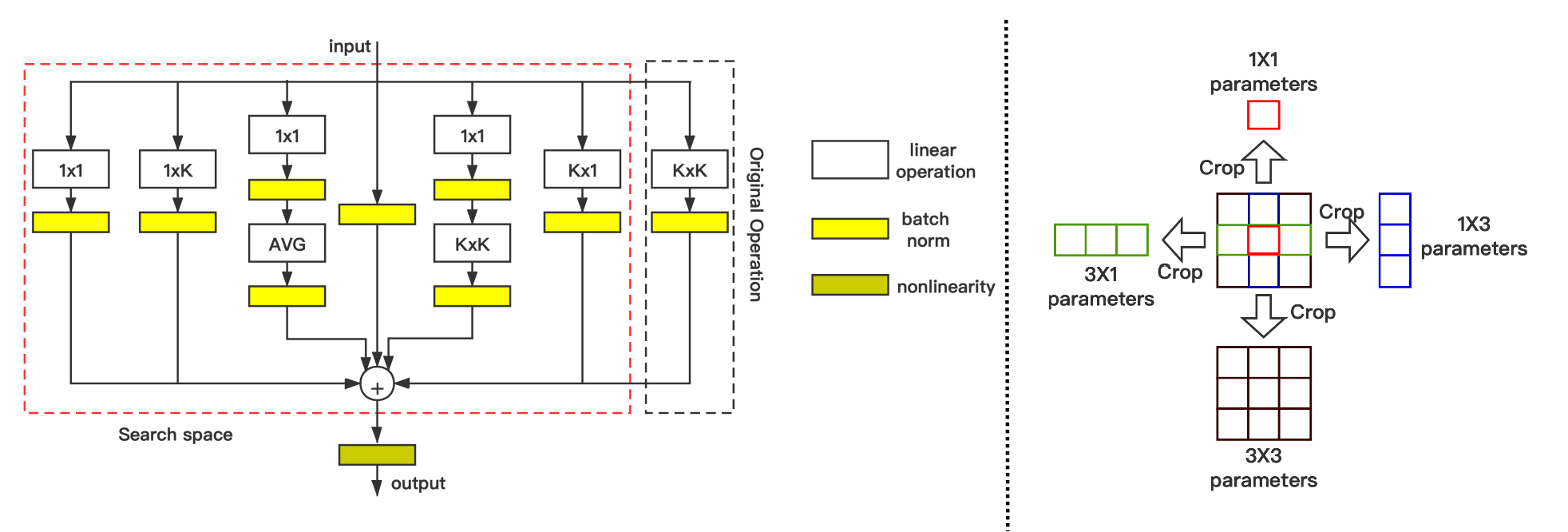}
    \caption{\textbf{Left:}The design of new search space based on several structural re-parameterization work\cite{ding2021repvgg,ding2021diverse,ding2019acnet}. Each block can preserve an arbitrary number of branches and is independent of other blocks. \textbf{Right}: The parameters of each branch are inherited from the original operation by center cropping.}
    \label{fig:search_space}
\end{figure*}
Inspired by BigNAS\cite{yu2020bignas} and slimmable networks\cite{yu2018slimmable}, we also share network parameters across the same block. For any branch that needs a convolutional operation, the weights of this branch can be inherited from the $K\times K$ convolutional operation in the main branch(see the right part of Figure \ref{fig:search_space}). We represent its weights as
\begin{equation}
  \begin{array}{lr}
    F^{H,W} = F^{K,K}_{:,:,LH:RH,LW:RW} \\
    LH = [K/2] - [H/2]\\
    RH = [K/2] + [H/2]\\
    LW = [K/2] - [W/2]\\
    RW = [K/2] + [W/2]
  \end{array}
\end{equation}
where $[.]$ is the floor operation, $H,W$ and $K,K$ denotes the kernel size of the inherited branch and main branch, respectively. Equipped with weight sharing, the supernet can get faster convergence, meanwhile, the ranking of branches can be evaluated precisely. We discuss the detail in \textbf{Ablation Study}.

\subsection{Searching for Rep Blocks}
\label{subsec:fitness}
To overcome the incorrect architecture ratings, an elegant solution is the one-stage differentiable NAS. The one-stage differentiable NAS is expected to simultaneously optimize the architecture parameter $\alpha$ and its network parameter $\omega$ by differentiable way, which can be formulated as the following optimization:
\begin{equation}
\label{optimization}
  \alpha^*, \omega^* = \mathop{argmin}\limits_{\alpha, \omega} L(\omega, \alpha; D_{train})
\end{equation}
where $L(\omega, \alpha; D_{train}) = E_{(x,y) \in D_train}[CE(y', y)]$ denotes the loss function computed in a specified training dataset. 

Many previous NAS works are designed to search on DARTs search space. With the heuristic rule, each edge only can preserve one branch. Therefore, in prior differentiable NAS work\cite{liu2018darts,xie2018snas}, the parameters probability $\alpha_{i,j}$ of the $j$th branch in the $i$th block($N$ branches) can be written as
\begin{equation}
\label{softmax}
  P(\alpha_{i,j}) = \frac{exp(\alpha_{i,j})}{\sum_{k=1}^{N}exp(\alpha_{i,k})}
\end{equation}
Though the Eq.(\ref{optimization}) can be optimized with gradient descent as most neural network training\cite{liu2018darts}, it would suffer from the huge performance gap between supernet and its child network\cite{chen2019progressive}. Instead, a continuous and differentialble reparameterization trick, $Gumbel\ softmax$\cite{jang2016categorical}, is used in NAS approaches\cite{dong2019searching,xie2018snas}. With $Gumbel$ random variable, Eq.(\ref{softmax}) can be rewritten as
\begin{equation}
\label{gumbel-softmax}
  Z_{i,j} = \frac{exp((\alpha_{i,j}+G_{i,j})/\lambda)}{\sum_{k=1}^{N}exp((\alpha_{i,k}+G_{i,j})/\lambda))}
\end{equation}
where $G_{i,j} = -log(-log(U_{i,j})$ and $U_{i,j}$ is a uniform random variable. $Z(\alpha_{i,j})$ can be approximated as a one-hot vector if temperature $\lambda -> 0$. This relaxation realizes the discretization of probability distribution.
In multi-branch search space proposed in \textbf{Rep Search Space Design}, each block can preserve an arbitrary branches so that we cannot directly obtain probability $\alpha$ as Eq.(\ref{softmax}) or Eq.(\ref{gumbel-softmax}). To fit the new space, we can regard whether each branch is retained or not as a binary classification. Hence, the discretization probability of the $j$th branch in the $i$th block can be given as
\begin{equation}
\label{bi-gumbel-softmax-0}
  Z^0_{i,j} = \frac{exp((\alpha^0_{i,j}+G^0_{i,j})/\lambda_{i,j})}{exp((\alpha^0_{i,j}+G^0_{i,j})/\lambda_{i,j})+exp((\alpha^1_{i,j}+G^1_{i,j})/\lambda_{i,j})}
\end{equation}
\begin{equation}
\label{bi-gumbel-softmax-1}
  Z^1_{i,j} = \frac{exp((\alpha^1_{i,j}+G^1_{i,j})/\lambda_{i,j})}{exp((\alpha^0_{i,j}+G^0_{i,j})/\lambda_{i,j})+exp((\alpha^1_{i,j}+G^1_{i,j})/\lambda_{i,j})}
\end{equation}
where $^0$ and $^1$ represent preserve and prune this branch, respectively. Different from Eq.(\ref{gumbel-softmax}), each branch in the new space is independent. Thus, the temperature $\lambda$ can be set differently according to requirements. Furthermore, we can combination the Eq.(\ref{bi-gumbel-softmax-0}) and Eq.(\ref{bi-gumbel-softmax-1}) into a sigmoid mode
\begin{equation}
\label{sigmoid}
             \begin{array}{lr}
             Z_{i,j} = f(\alpha_{i,j}) \\
              = \frac{1}{1 + exp((\alpha^0_{i,j}+G^0_{i,j} - \alpha^1_{i,j} - G^1_{i,j})/\lambda_{i,j})} = \frac{1}{1 + exp((\alpha_{i,j}+\zeta_{i,j})/\lambda_{i,j})}
             \end{array}
\end{equation}
where $\zeta \sim~Logistic(0, 1)$. We only need to optimize $\alpha$, instead of $\alpha^0$ and $\alpha^1$, through grident descent. It grident can be given as
\begin{equation}
\label{gradient}
  \frac{\partial L}{\partial \alpha_{i,j}} = \frac{\partial L}{\partial x_{i}}O_{i,j}^Tf(\alpha_{i,j})(1-f(\alpha_{i,j}))/\lambda_{i,j} 
\end{equation} 
where $x_i$ and $O_{i,j}$ denotes the output of the $i$th block and the $j$th branch, respectively. In each training iteration, we can firstly compute a threshold $s$ according to the global ranking. Subsequently, the branches whose ranking is below the $s$ will be pruned out and do not participate in current forward or backward. Thanks to the independence of each branch, we can easily control the activation of each branch through its temperature $\lambda$. 
\begin{equation}
\label{branch activation}
  \left\{
             \begin{array}{lr}
             \mathop{lim}\limits_{\lambda_{i,j} \rightarrow 0^+} Z_{i,j}=0, \qquad if\ R_{i,j}<0 \ and\ rank(R_{i,j})<s \\
             \mathop{lim}\limits_{\lambda_{i,j} \rightarrow 0^-} Z_{i,j}=0, \qquad if\ R_{i,j}>0\ and\ rank(R_{i,j})<s \\
             \mathop{lim}\limits_{\lambda_{i,j} \rightarrow 0^-} Z_{i,j}=1, \qquad if\ R_{i,j}<0\ and\ rank(R_{i,j})>s \\
             \mathop{lim}\limits_{\lambda_{i,j} \rightarrow 0^+} Z_{i,j}=1, \qquad if\ R_{i,j}>0\ and\ rank(R_{i,j})>s \\
             \end{array}
\right.
\end{equation}
where $R_{i,j}=\alpha_{i,j}+\zeta_{i,j}$ is a random variable. $rank(R_{i,j})$ denotes the global ranking of the $(i,j)$ branch. 
\begin{table*}[t]
\caption{Experimental configurations.} 
\label{exp_config}
\begin{center}
\begin{tabular}[c]{c|c|c|c|c|c|c}
\toprule[2pt]
Dataset&Arch.&Epochs&Batch size& Init LR& Weight decay& Data augmentation\\
\toprule[2pt]
CIFAR-10&VGG-16&600&128&0.1&$1\times10^{-4}$&same as \cite{ding2021diverse}\\
ImageNet&ODBB-A0/A1/A2/B1/B2/ResNet-18&150&256&0.1&$1\times10^{-4}$&same as \cite{ding2021repvgg}\\
ImageNet&ODBB-B3/ResNet-101&240&256&0.1&$1\times10^{-4}$&same as \cite{ding2021repvgg}\\
COCO&ResNet18/ODBB-A0&140&114&5e-4&0&same as \cite{zhou2019objects}\\
COCO&ResNet101/ODBB-B3&140&96&3.75e-4&0&same as \cite{zhou2019objects}\\
\bottomrule[2pt]
\end{tabular}
\end{center}
\end{table*}
In implementation, we sort the importance of each branch by Eq.(\ref{sigmoid}) and only keep the top-k branches for forward. The importance $Z_{i,j}$ of unpruned $(i, j)$ branch is convergence to $1$ by Eq.(\ref{branch activation}). Afterward, we simply multiply $Z_{i,j}$ to the output of $(i,j)$ unpruned branch, such that $\frac{\partial L}{\partial \alpha_{i,j}}$ can be obtained by the chain rule. The whole algorithm is shown in Alg\ref{alg:modified_dsnas}.
\begin{algorithm}[t]\small
\caption{The algorithm detail of RepNAS.}
\label{alg:modified_dsnas}
\begin{algorithmic} 
\REQUIRE unpruned network, branch parameters $\bm{\theta}$, arch parameters $\bm{\alpha}$ and branch number constraint $C$
\STATE Initialize $\bm{\theta}$, $\bm{\alpha}$
\WHILE{not converged} 
\STATE Calculate the importance of each branch $\bm{Z}$ by Eq.(\ref{sigmoid}) where $\lambda = 1$
\STATE Discretize $Z$ by Eq.(\ref{branch activation}) to met $C$
\STATE Sample branches where $\bm{Z=1}$
\STATE Get a batch from data and forward to get $L$, multiply $\bm{Z}$ after each feature map $X$
\STATE Backward $L$ to both $\bm{\theta}$ and $\bm{\alpha}$
\STATE Update $\bm{\theta}$ with $\frac{\partial L}{\partial \bm{\theta}}$, update $\bm{\alpha}$ with $\frac{\partial L}{\partial \bm{\alpha}}$
\ENDWHILE
\end{algorithmic}
\end{algorithm}
\begin{table*}[t]
\caption{RepNAS performance on ImageNet with comparisons to other NAS methods. We group the models according to their Top-1 accuracy. The inference time is tested on a NVIDIA Tesla V100 GPU/NVIDIA Xavier with a batch size of 128/1, full precision(fp32).}
\label{imagenet_nas}
\begin{center}
\begin{tabular}[c]{c|c|c|c|c|c|c}
\toprule[2pt]
Model&Parameters(M)&FLOPs(B)&Inference(s)&Search Space&Search+Retrain Cost(GPU Days)&Top-1(\%)\\
\toprule[2pt]
\textbf{ODBB(A1)}&12.78&2.4&\textbf{0.031/0.028}&\textbf{Rep(Ours)}&24+0&75.24\\
DARTS&4.9&0.59&0.067/0.178&DARTS&1+24&73.1\\
SNAS&4.3&0.52&0.063/0.167&DARTS&1.5+24&72.7\\
NASNet-A&5.3&0.56&0.078/0.195&DARTS&1800+24&74.0\\
AmoebaNet-B&4.9&0.56&0.218&DARTS&3150+24&74.0\\
\hline
\textbf{ODBB(A2)}&25.49&5.1&\textbf{0.038/0.031}&\textbf{Rep(Ours)}&24+0&76.86\\
SPOS&3.5&0.32&0.061/0.053&ShuffleNet&24+24&74.0\\
BigNASModel-S&4.5&0.24&0.045/0.049&MobileNet&40+0&76.5\\
Once-For-All&4.4&0.23&0.051/0.057&MobileNet&40+24&76.4\\
\hline
\textbf{ODBB(B3)}&110.96&26.2&\textbf{0.107/0.106}&\textbf{Rep(Ours)}&50+0&80.97\\
BigNASModel-L&9.5&1.1&0.186/0.265&MobileNet&80+0&80.9\\
\bottomrule[2pt]
\end{tabular}
\end{center}
\end{table*}
\section{Experiments} 
\label{sec:experiments}
We first compare ODBB with baseline, random search, DBB\cite{ding2021diverse} and ACB\cite{ding2019acnet} on CIFAR-10 for a quick sanity check. To further demonstrate the effectiveness of ODBB with various model size, experiments on a large-scale dataset, ImageNet1k\cite{krizhevsky2012imagenet} are conducted. At last, the impact of weight sharing and the effect of the branch number constraints on model performance is given in the ablation study.
\subsection{Quick Sanity Check on CIFAR-10} 
\label{sub:quick_check}
We use VGG-16\cite{simonyan2014very} as the benchmark architecture. The convolutional operations in the benchmark architecture are replaced by ODBB, DBB and ACB, respectively. For a fair comparison, the data augmentation techniques and training hyper-parameter setting are followed with DBB and ACB which can be given in Table \ref{exp_config}. To optimize architecture parameter $\alpha$, simultaneously, we use Adam\cite{kingma2015adam} optimizer with $0.0001$ learning rate and $(0.5, 0.999)$ betas. One of the indicators to evaluate the effectiveness of a NAS approach is whether the search result exceeds the result of the random search. To produce the random search result, 25 architectures are randomly sampled from the supernet. We train each of them for 100 epochs and pick up the best one for an entire 600-epoch re-training.
The comparison results are shown in Table \ref{cifar10}. ODBB can improve VGG-16 on CIFAR-10 by 0.85\% and surpass the DBB and ACB. The architecture generated by random search also can slightly improve the performance of the benchmark model but fall behind the ODBB by 0.77\%, which demonstrates the effectiveness of our proposed NAS algorithm. 
\begin{table}
\caption{Top-1 Acc. of VGG-16 baseline, random search, DBB, ACB and ODBB on CIFAR-10.} 
\label{cifar10}
\begin{center}
\begin{tabular}[c]{c|c|c}
\toprule[2pt]
Method&Top-1 Acc.(\%)&Improvement(\%)\\
\toprule[2pt]
Baseline&93.83&-\\
Random search&93.91&0.08\\
DBB\cite{ding2021diverse}&94.45&0.62\\
ACB\cite{ding2019acnet}&94.13&0.30\\
\textbf{ODBB(Ours)}&\textbf{94.68}&0.85\\
\bottomrule[2pt]
\end{tabular}
\end{center}
\end{table}
\subsection{Performance improvements on ImageNet} 
\label{sub:performance_improvements_on_imagenet}
To reveal the generalization ability of our method, we then search for a series of models on ImageNet-1K which comprises 1.28M images for training and 50K for validation from 1000 classes. RepVgg-series\cite{simonyan2014very} are used as benchmark architectures. We replace each original RepVGG block by designed block in \textbf{Rep Search Space Design}. For a fair comparison, the total branch number of all ODBB models is limited to be equal to RepVGGs. The training strategies of each network are listed in Table \ref{exp_config}. We also use Adam optimizer with $0.0001$ learning rate and $(0.5, 0.999)$ betas to optimize $\alpha$.  We will discuss the impact of the branch number on performance in \textbf{Ablation Study}.

\begin{table}[t]
\caption{Top-1 Acc. of several benchmark architectures on ImageNet. the structural re-parameterization techniques of RepVGG do not include in ACB and DBB. Thus, we only compare the performance with the original RepVGG.}
\label{imagenet}
\begin{center}
\begin{tabular}[c]{c|c|c|c|c}
\toprule[2pt]
Arch.&Param.(M)&Bran.&Ori.(\%)&ODBB(\%)\\
\toprule[2pt]
A0&8.3&66&72.41&72.96\\
A1&12.78&66&74.46&75.24\\
A2&25.49&66&76.48&76.86\\
B1&14.33&84&78.37&78.61\\
B2&80.31&84&78.78&79.10\\
B3&110.96&84&80.52&80.97\\
\bottomrule[2pt]
\end{tabular}
\end{center}
\end{table}
Results are summarized in Table \ref{imagenet}. RepVGG is stacked with several multi-branch blocks that contain a $1\times 1\ Conv$, a $3\times 3\ Conv$ and a skip connection and also can be fused into a single path in inference. We search for more powerful architectures for blocks in RepVGG. For RepVGG-A0-A2 that has 22 layers, ODBB can achieve 0.55\%, 0.38\% and 0.24\% better accuracy than original baselines, respectively. To our best knowledge, RepVGG-B3 with ODBB refreshes the performance for plain models from 80.52\% to 80.97\%. Noting that ACB, DBB and RepVGG Block are special cases of our proposed search space, our proposed NAS really can search the optimal architecture beyond human-defined ones.
\subsection{Comparison with Other NAS} 
\label{sub:comparison_with_other_nas}
We compare ODBB-series models with other models searched from DARTS and mobile search space on ImageNet-1K. We verify the real inference time on two different hardware platforms, including GPU(NVIDIA Tesla V100) and embedded device(NVIDIA Xavier). Table \ref{imagenet_nas} presents the results and shows that: 1) ODBB-series models searched from Rep search space consistently outperform other networks searched from other search space with lower inference time since ODBB-series networks are only stacked by 3x3 convolutions and have no branches. 2) RepNAS can directly search on ImageNet-1k and combine the train and search, which brings high performance and low GPU days cost. We show the architectures of searched ODBB-series in the Appendix.
\subsection{transferability} 
\label{sub:transferability}
To verify the transferability of searched models on the other computer version task, object detection experiments are conducted on MS COCO dataset\cite{lin2014microsoft} with CenterNet framework\cite{zhou2019objects}. The detailed implementation is following \cite{zhou2019objects}, ImageNet-1k pretrained backbones with three up-convolutional networks are finetuned on COCO. The input resolution is $512 \times 512$. We use Adam to optimize the overall parameters. Specifically, the baseline backbone of CenterNet is ResNet-series and we replace it with our searched ODBB-series. All backbone networks are pretrained on ImageNet-1K with train schedule in Table \ref{exp_config}. After finetuning on COCO, ODBB-series will be fused into one path for fast inference. The results are shown in Table \ref{detection}. Both in slight and heavy models, ODBB-A0 and ODBB-B3 surpass ResNet-18 and ResNet-101 by 3.3\% and 2.7\% AP with comparable inference speed, respectively, which demonstrates our searched models have outstanding performance in other computer version tasks.

\subsection{Ablation Study} 
\label{sub:ablation_study}
\begin{table}[t]
\caption{Object detection on COCO validation.}
\label{detection}
\begin{center}
\begin{tabular}[c]{c|c|c}
\toprule[2pt]
Backbone& COCO AP(\%) & Inference(s)\\
\toprule[2pt]
ResNet-18& 28.1 & 0.065\\
ResNet-101& 34.6 & 0.141\\
\hline
ODBB(A0)& 31.4 & 0.048\\
ODBB(B3)& 37.3 & 0.148\\
\bottomrule[2pt]
\end{tabular}
\end{center}
\end{table}
\textbf{Training Under Branch Number Constraint.} Training multi-branch models requires linearly increased GPU memory to save the middle feature maps for forward and backward. How to memory-friendly search and train multi-branch model is significant for the application of structural re-parameterization techniques. For simplicity, we reduce the branch number of the network to meet the various memory constraints by pushing temperature $\lambda$ of low-rank branches to $0^-$. To prove the efficiency and effectiveness over other human-designed blocks, we also train DBB and ACB on benchmark RepVGG-A0 by replacing the original RepVGG block, respectively. All experiments in this subsection are conducted on RepVGG-A0 with the same training set. 
\begin{figure}[t]
    \centering
    \includegraphics[width=3in]{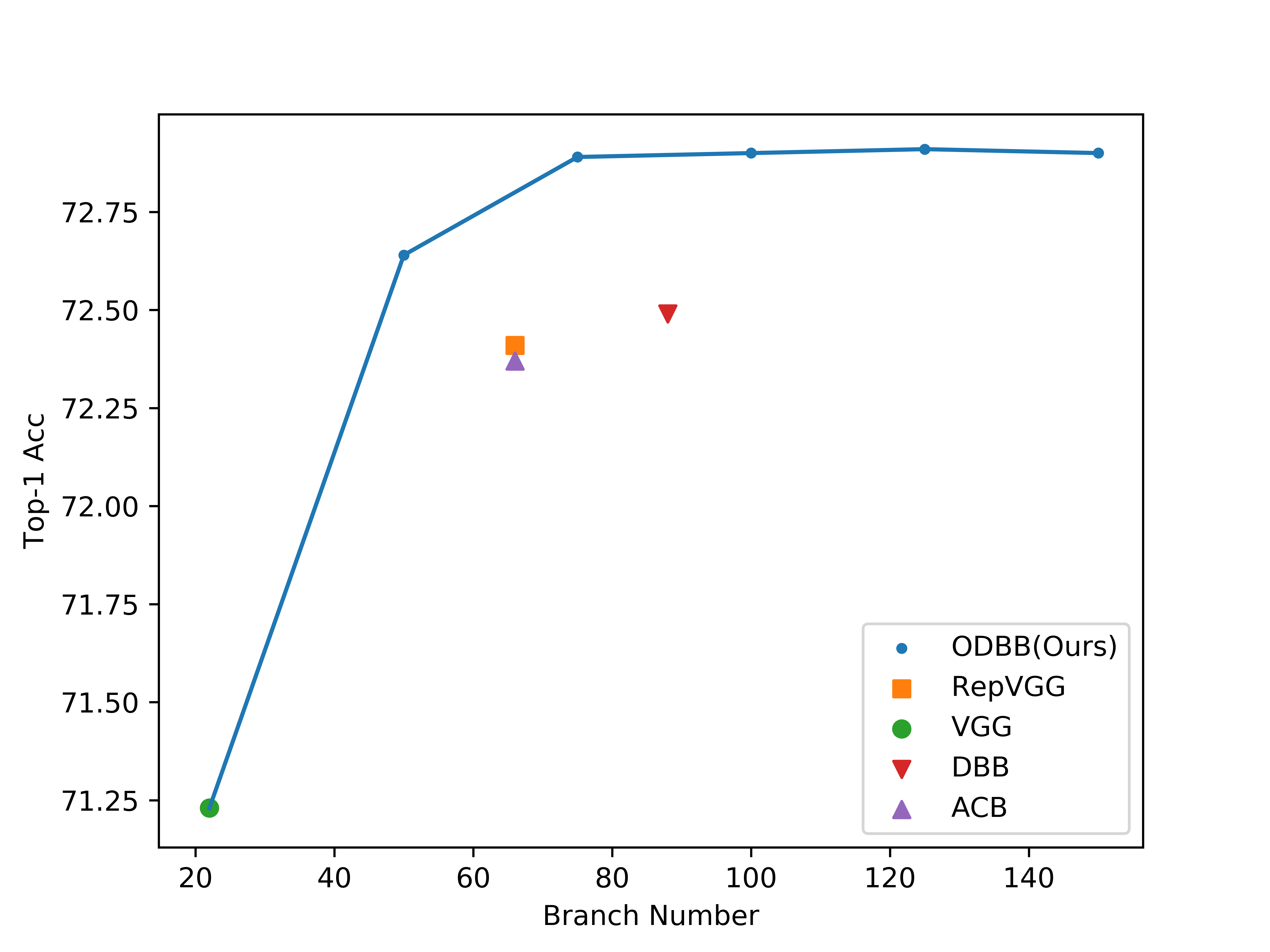}
    \caption{The tradeoff curve of top-1 Acc. and branch number. We compare ODBB with other manual block architectures(i.e. original RepVGG block, DDB, ACB) and the single path model that only has one $3\times3\ Conv$ in each block.}
    \label{fig:path}
\end{figure}
\begin{figure}[t]
    \centering
    \includegraphics[width=3in]{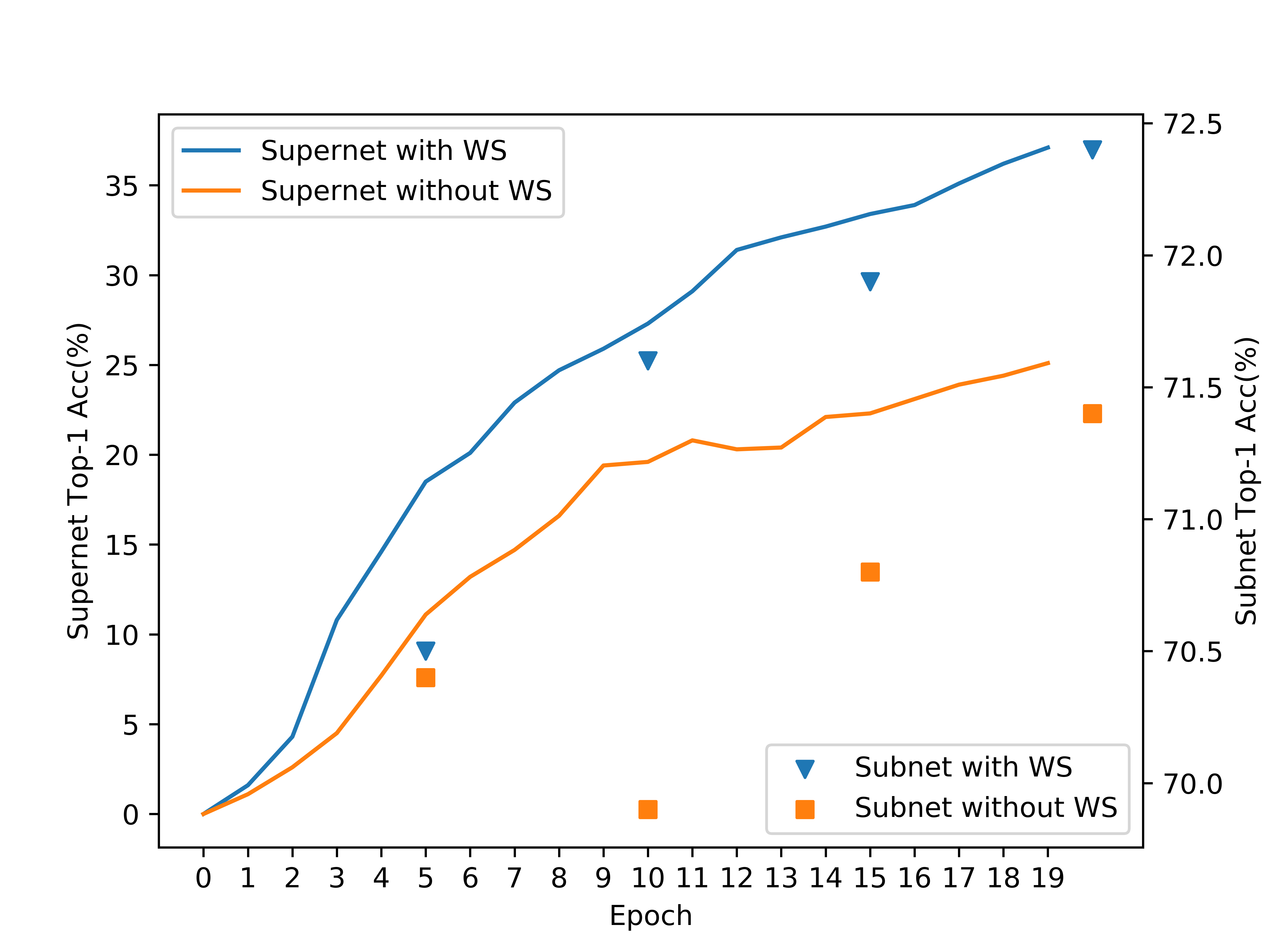}
    \caption{Top-1 Acc on ImageNet of supernet and sampled high-performing subnet.}
    \label{fig:ws}
\end{figure}
The results shown in Figure \ref{fig:path} reveal that the performance of ODBB can surpass other blocks with fewer branches. For instance, ODBB only containing 50 branches has higher top-1 Acc. over the original RepVGG block(66 branches), DBB(88 branches) and ACB(66 branches). Besides, we found that the ODBB with 75 branches is enough to obtain the same results as the ODBB with more branches.

\textbf{Efficacy of Weight Sharing.} We train and search ODBB(A0) with the following two methods: 1) The weights of the different branches in the same block are shared. 2) The weights of the different branches are independently updated. Figure \ref{fig:ws} presents the comparisons on ImageNet-1k with ODBB(A0). It is apparent that the ODBB(A0) updated with weight sharing gets faster convergence than the one that is updated independently. Besides, we sample the high-performing subnet in every five training epochs according to the ranking of the architecture parameter $\alpha$. Each sampled model is trained from scratch with the scheme in Table \ref{exp_config}. Shown as Figure \ref{fig:ws}, the performance of subnet sampled from weight sharing supernet has more stable and higher accuracy in each period. This phenomenon illustrates the weight-sharing supernet serves as a good indicator of ranking.

\section{Limitations}
The searched networks have the same channels and depth as RepVGG\cite{ding2021repvgg} networks which contain huge parameters. Therefore, less concerning the number of parameters prevents the deployment of our searched models from CPU devices. However, VGG-like models can be easily pruned by many existing fast channel pruning methods\cite{ding2020lossless,chin2020towards,he2017channel} for parameter reduction.

\section{Conclusion}
In this work, we first propose a new search space, Rep space, where each block is architecture-independent and can preserve arbitrary branches. Any subnet searched from Rep space has fast inference speed since it can be converted to a single path model in inference. To efficiently train the supernet, block-wisely weight sharing is used in supernet training. To fit the new search space, a new one-stage NAS method is presented. The optimal diverse branch blocks can be obtained without retraining. Extensive experiments demonstrate that the proposed RepNAS can search various sizes of multi-branch networks, named ODBB-series, which strongly outperform the previous NAS networks under a similar inference time. Moreover, Compared with other networks utilized Rep techniques, ODBB-series also achieve the state-of-the-art top-1 accuracy on ImageNet in various model size. In the feature work, we will consider adding the channel number of each block to search space, which can further boost the performance of RepNAS.

{\small
\bibliographystyle{ieee_fullname}
\bibliography{egbib}
}
\newpage
\section*{Appendix: Model Architecture}
The design of channels and depth of ODBB-series networks is following RepVGG\cite{ding2021repvgg}, we show it in Table \ref{channel_number_depth}.
\begin{table}[h]
\label{channel_number_depth}
\caption{The channel number and depth of each stage.}
\begin{center}
\begin{tabular}[c]{c|c|c}
\toprule[2pt]
Name& Layers of each stage & Channels of each stage\\
\toprule[2pt]
A0& 1, 2, 4, 14, 1& 64, 48, 96, 192, 1280\\
A1& 1, 2, 4, 14, 1& 64, 64, 128, 256, 1280\\
A2& 1, 2, 4, 14, 1& 64, 96, 192, 384, 1408\\
\hline
B1& 1, 4, 6, 16, 1& 64, 128, 256, 512, 2048\\
B3& 1, 4, 6, 16, 1& 64, 160, 320, 640, 2560\\
B3& 1, 4, 6, 16, 1& 64, 192, 384, 768, 2560\\
\bottomrule[2pt]
\end{tabular}
\end{center}
\end{table}
\label{appendixa}
\begin{figure}[htbp]
    \centering
    \includegraphics[width=3.5in]{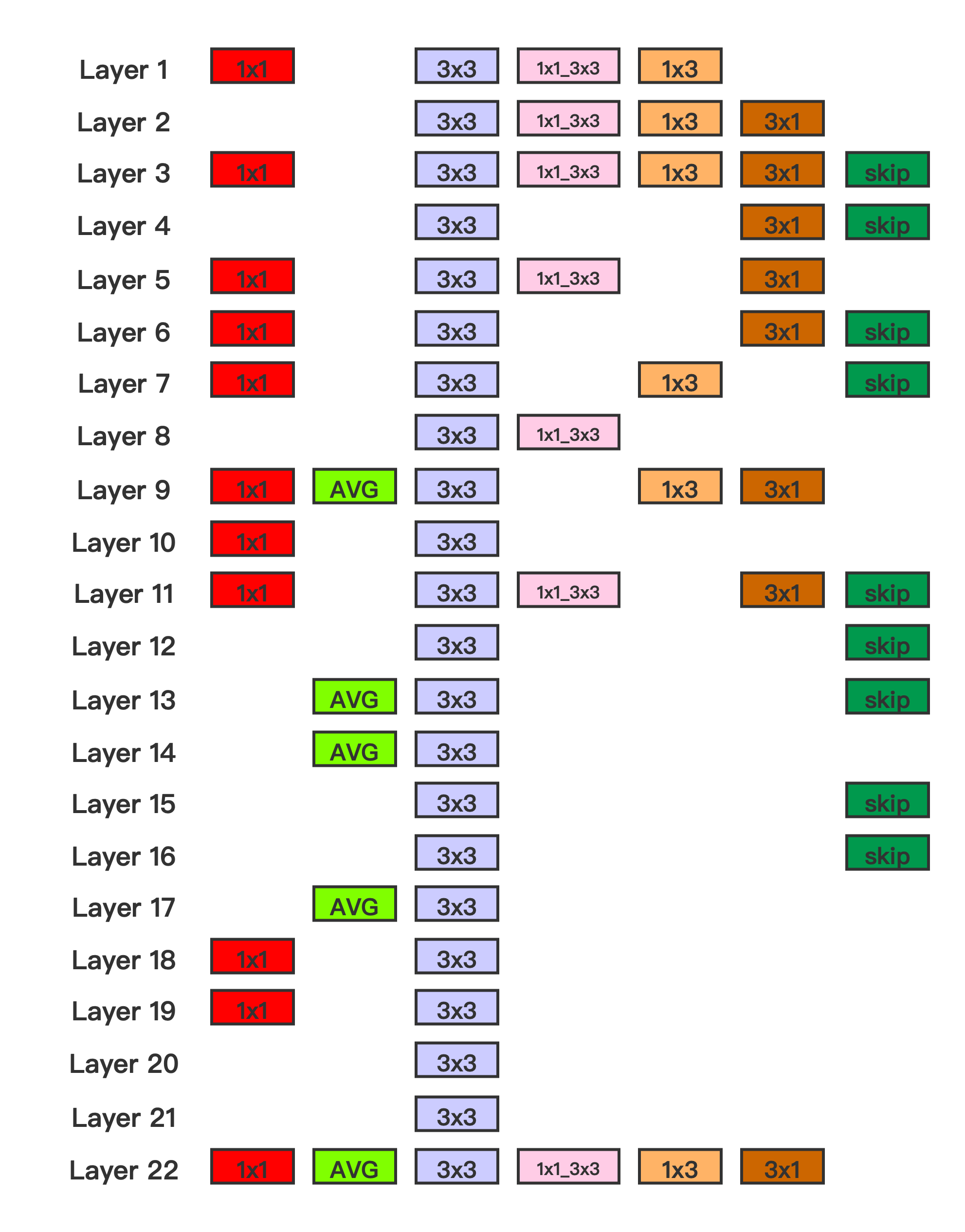}
    \caption{Searched result of ODBB-A0}
\end{figure}

\begin{figure}[htbp]
    \centering
    \includegraphics[width=3.5in]{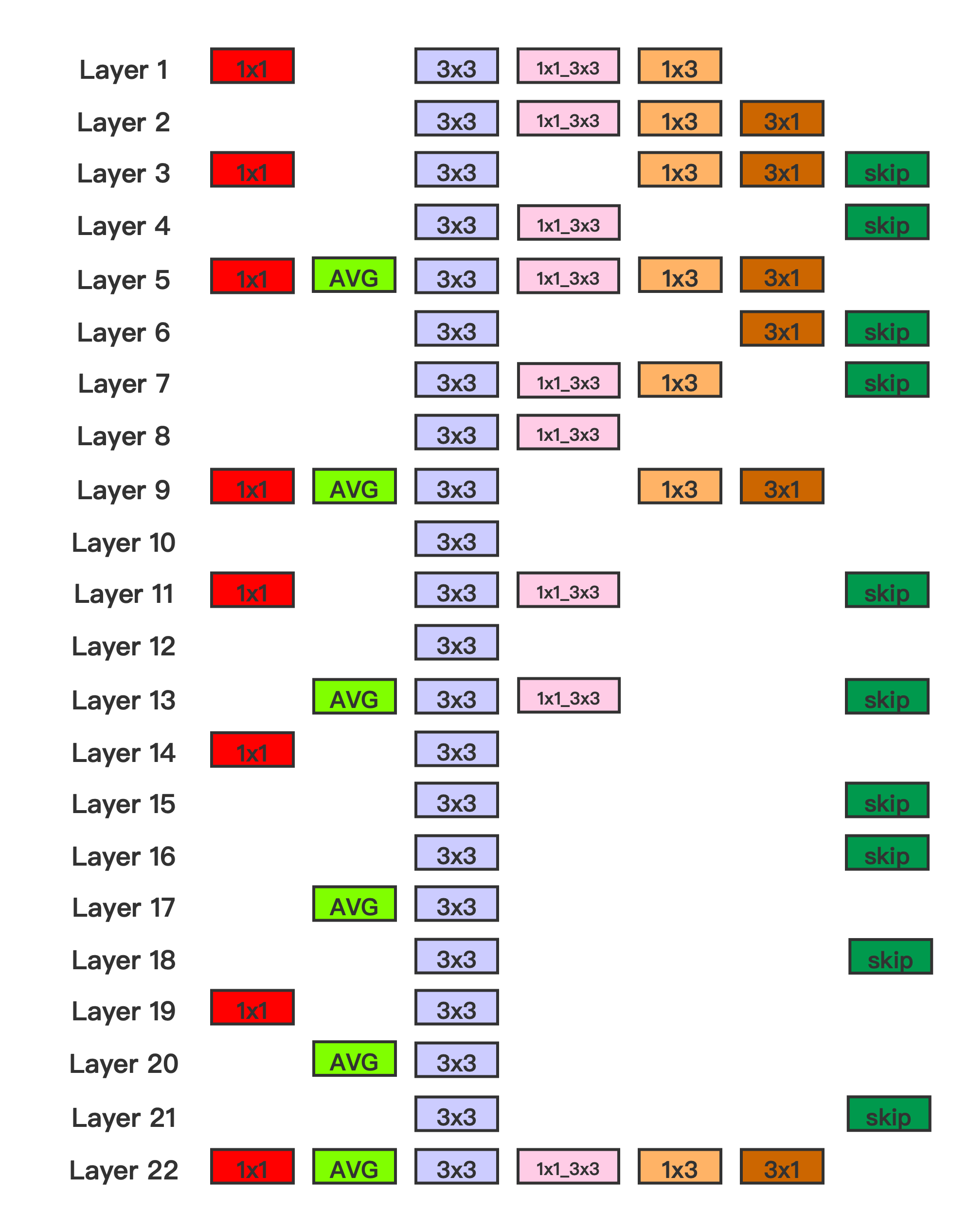}
    \caption{Searched result of ODBB-A1}
\end{figure}

\begin{figure}[htbp]
    \centering
    \includegraphics[width=3.5in]{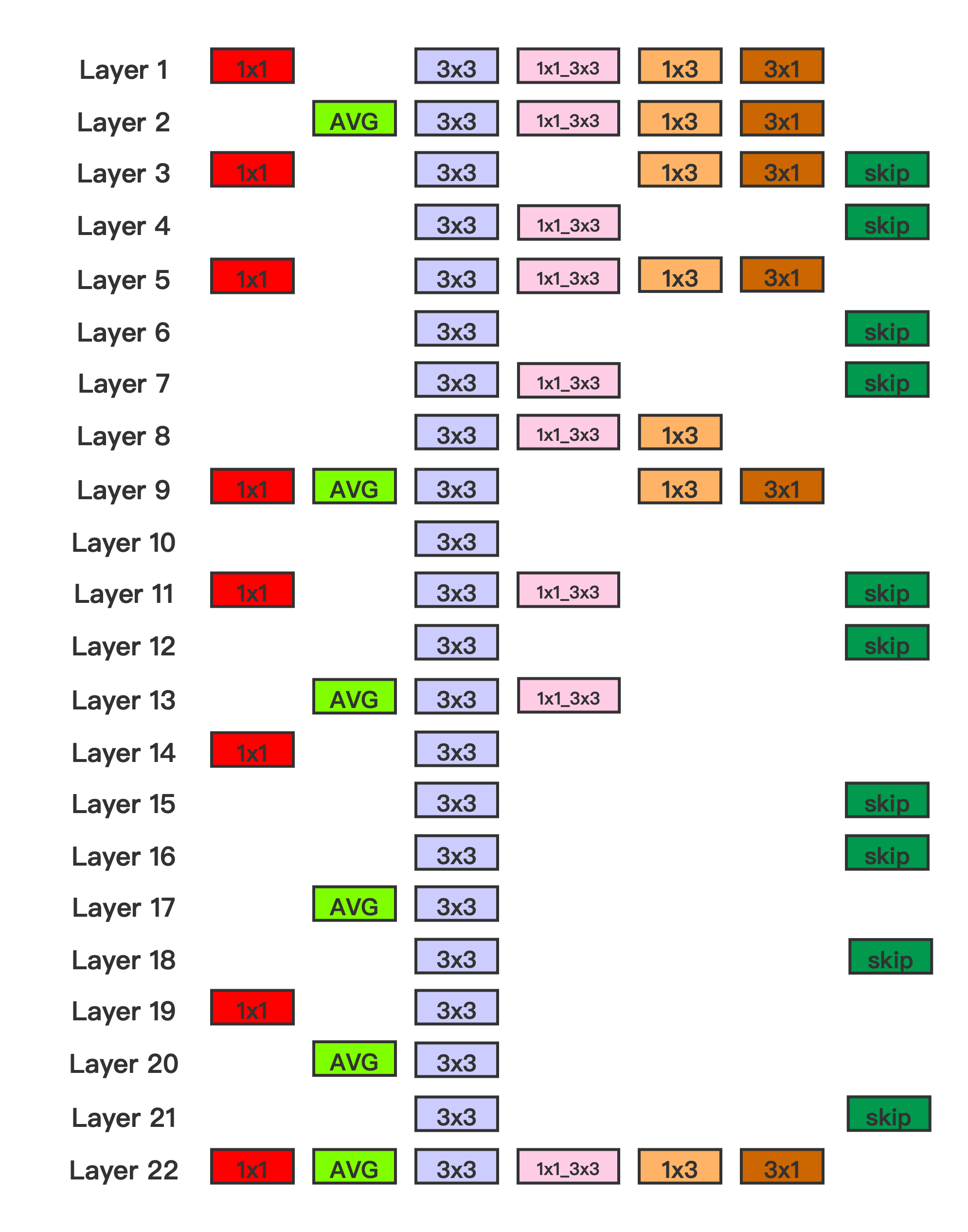}
    \caption{Searched result of ODBB-A2}
\end{figure}

\begin{figure}[htbp]
    \centering
    \includegraphics[width=3.5in]{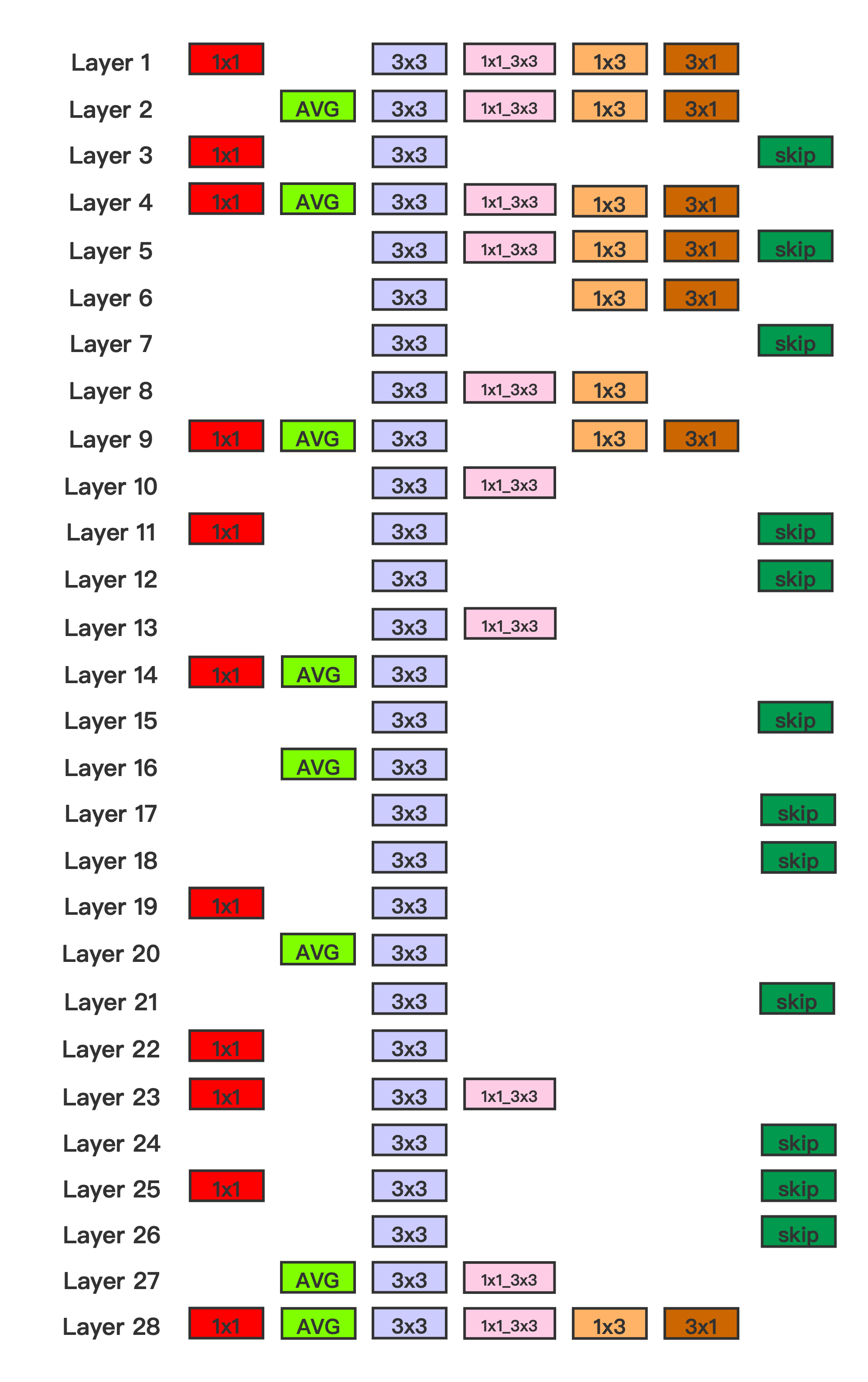}
    \caption{Searched result of ODBB-B1}
\end{figure}

\begin{figure}[htbp]
    \centering
    \includegraphics[width=3.5in]{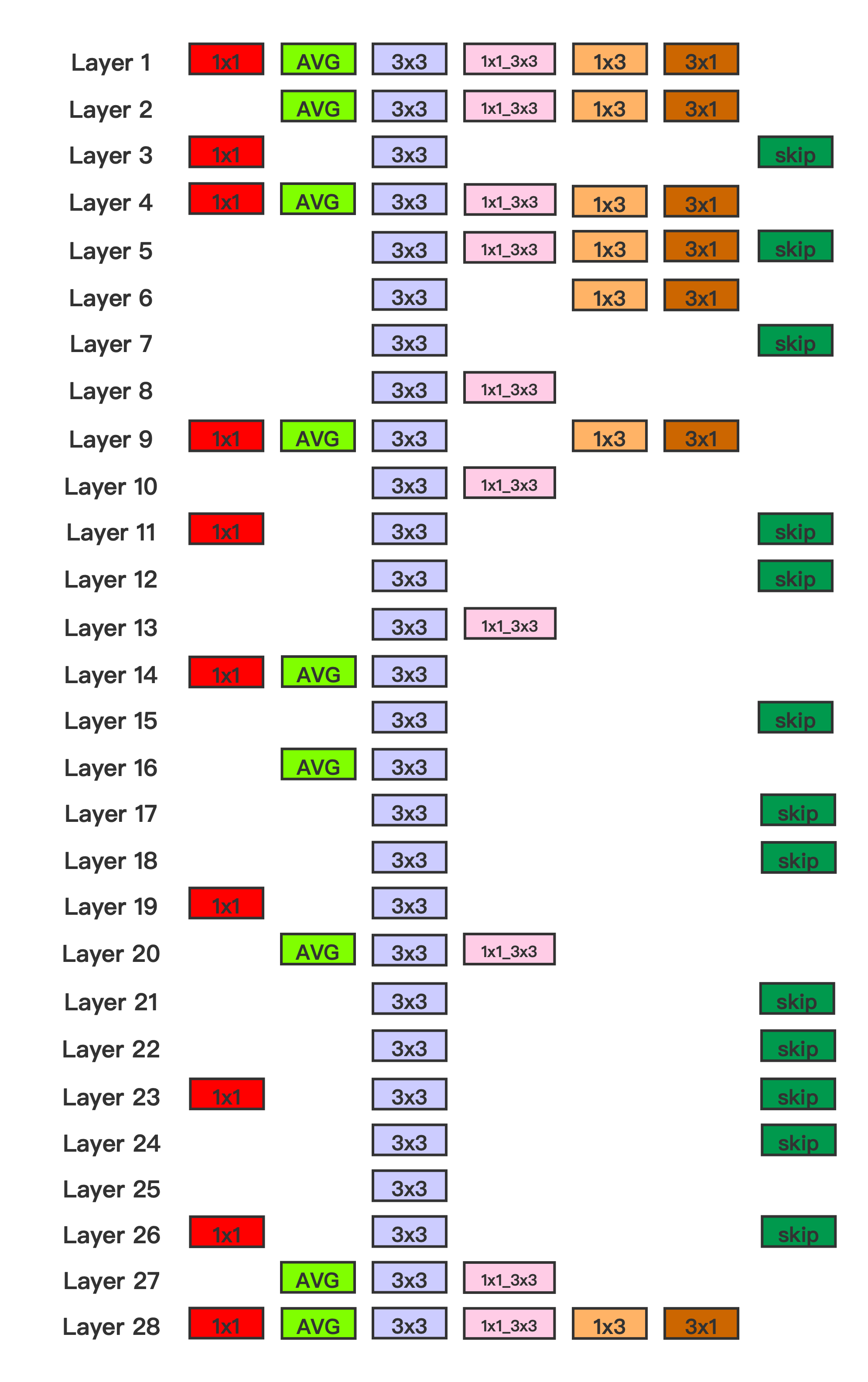}
    \caption{Searched result of ODBB-B2}
\end{figure}

\begin{figure}[htbp]
    \centering
    \includegraphics[width=3.5in]{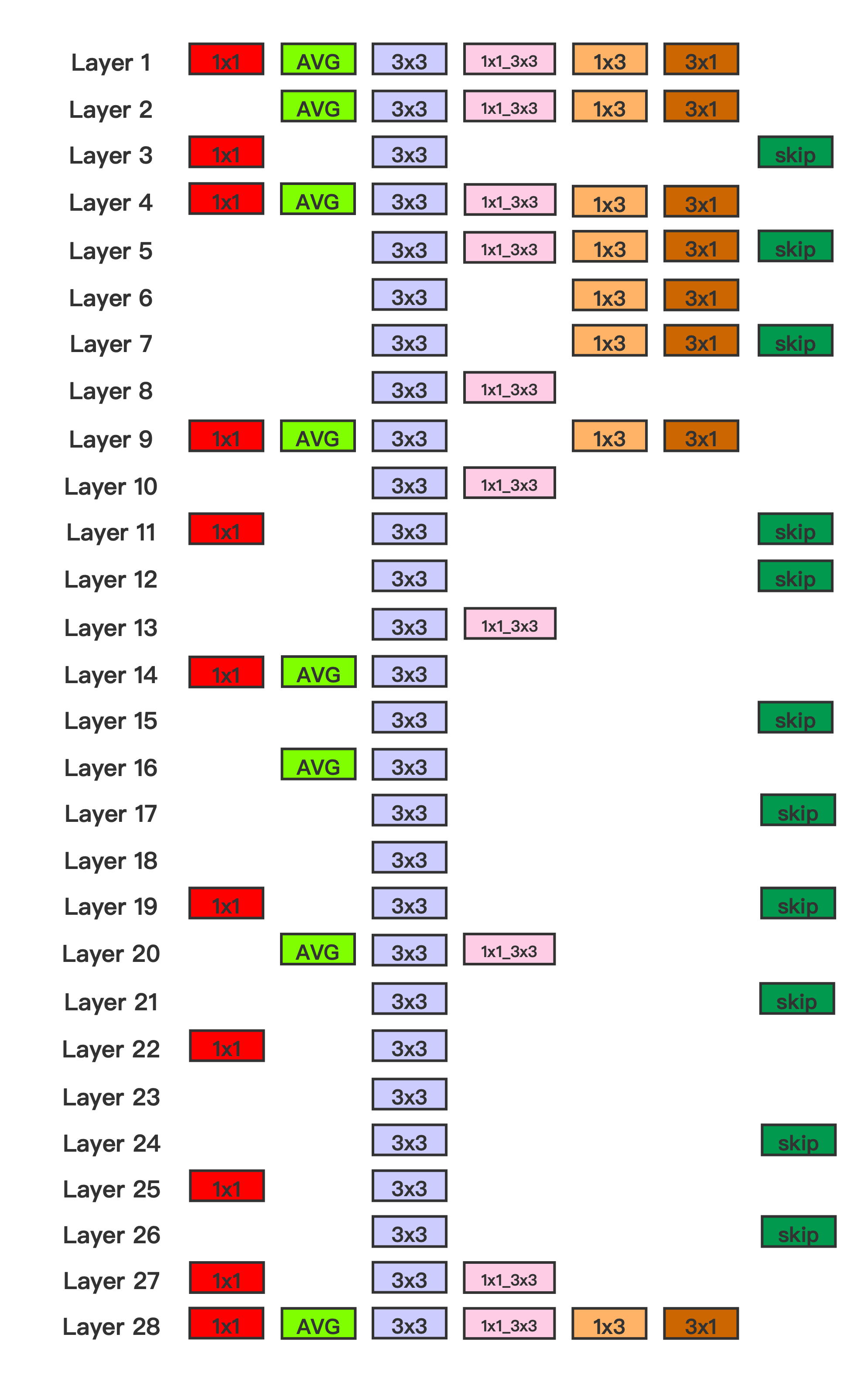}
    \caption{Searched result of ODBB-B3}
\end{figure}
\end{document}